\pdfoutput=1

\documentclass[11pt]{article}

\usepackage[table]{xcolor}
\usepackage[]{acl}
\usepackage{graphicx}
\graphicspath{ {./images/} }
\usepackage{caption}
\usepackage{subcaption}

\usepackage{enumitem}
\usepackage{array}

\usepackage{times}
\usepackage{latexsym}

\usepackage[T1]{fontenc}

\newif\ifvpversion
\vpversionfalse

\usepackage[utf8]{inputenc}

\usepackage{microtype}

\usepackage{cleveref}
\crefformat{section}{\S#2#1#3} 
\crefformat{subsection}{\S#2#1#3}
\crefformat{subsubsection}{\S#2#1#3}

\newcommand{\textinput}[1]{``#1''}

\newcommand{\etal}{{et al.\ }}
\newcommand{\aiao}{\textsc{aiao}}
\newcommand{\aido}{\textsc{aido}}

\title{Underspecification in Scene Description-to-Depiction Tasks}

\author{Ben Hutchinson \\
  Google Research, Australia \\
  \texttt{\small benhutch@google.com} \\\And
  Jason Baldridge \\
  Google Research, USA \\
  \texttt{\small jasonbaldridge@google.com} \\\And
  Vinodkumar Prabhakaran \\
  Google Research, USA \\
  \texttt{\small vinodkpg@google.com} \\}

\begin{document}
\maketitle
\begin{abstract}
Questions regarding implicitness, ambiguity and underspecification are crucial for understanding the task validity and ethical concerns of multimodal image+text systems, yet have received little attention to date. 
This position paper maps out a conceptual framework to address
this gap,
focusing on systems which generate images depicting scenes from scene descriptions. In doing so, we account for how texts and images convey meaning differently. 
We outline a set of core challenges concerning textual and visual ambiguity, as well as risks that may be amplified by ambiguous and underspecified elements.
We propose and discuss strategies for addressing these challenges, including generating visually ambiguous images, and generating a set of diverse images.
\end{abstract}


\section{Introduction}

The classic Grounding Problem in AI asks \emph{how is it that language can be interpreted as referring to things in the world?} It has been argued that demonstrating natural language understanding requires mapping text to something that is non-text and that functions as a model of meaning \cite[e.g.,][]{bender2020climbing}. In this view, multimodal models that relate images and language have an important role in pursuing contextualized language understanding. Indeed, joint modeling of linguistic and visual signals has been argued to play a critical role in progress towards this ultimate goal, as precursors to modeling relationships between language and the social and physical worlds
\cite{bisk2020experience}.

Recent {text-to-image} \textit{generation} systems have demonstrated impressive capabilities \cite{Zhang_2021_CVPR,Ramesh21dalle,ding2021cogview,glide2022,makeascene2022,ramesh2022hierarchical, saharia2022photorealistic,ramesh2022hierarchical,yu2022scaling}.
These employ deep learning methods such as generative adversarial networks
\cite{goodfellow2014generative}, neural discrete representation learning \cite{Oord17vqvae} combined with auto-regressive models \cite{brown2020gpt3}, and diffusion models \cite{pmlr-v37-sohl-dickstein15}, trained on large datasets of images and aligned texts \cite{radford2021learning,pmlr-v139-jia21b}.

\begin{figure}
    \centering
         \begin{subfigure}[b]{0.19\textwidth}
         \centering
         \fbox{\fbox{\includegraphics[width=0.9\textwidth]{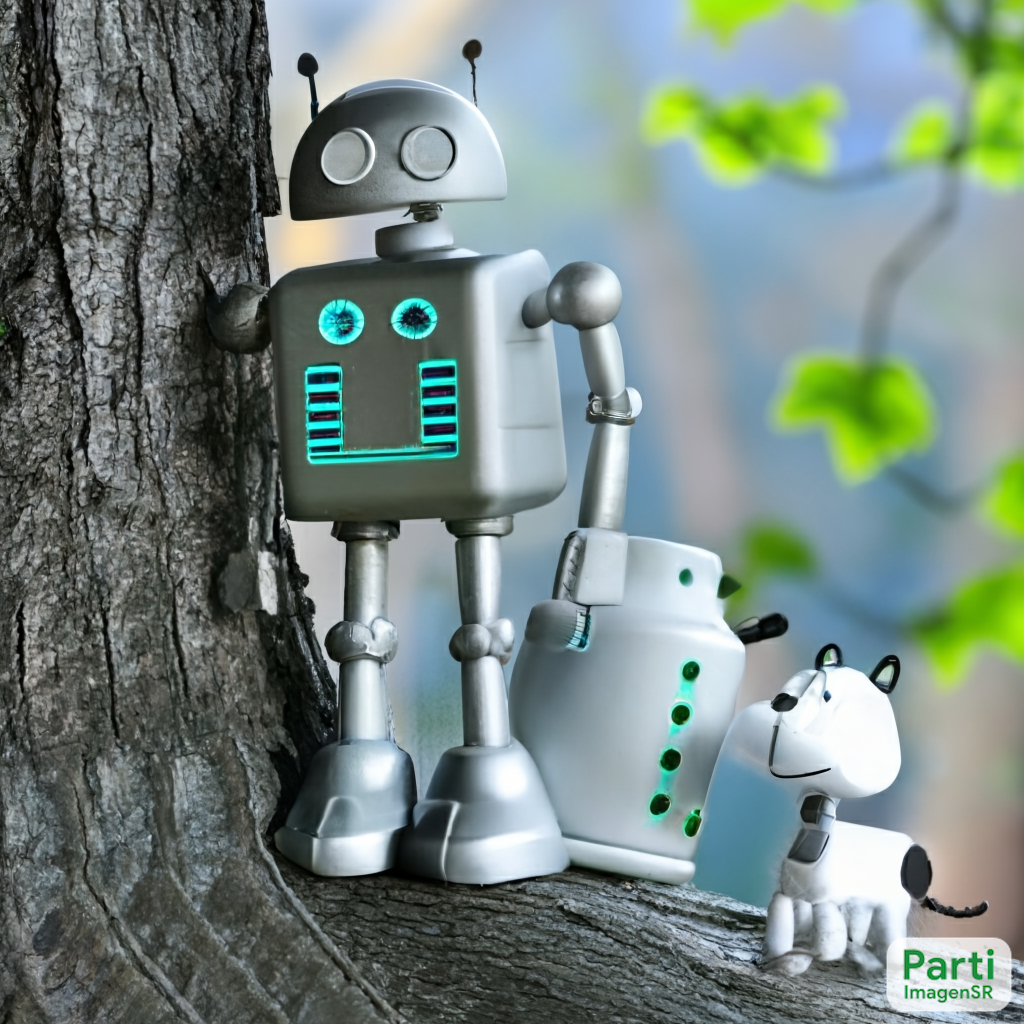}}}
     \end{subfigure}
     \quad \quad
     \begin{subfigure}[b]{0.19\textwidth}
         \centering
         \fbox{\fbox{\includegraphics[width=0.9\textwidth]{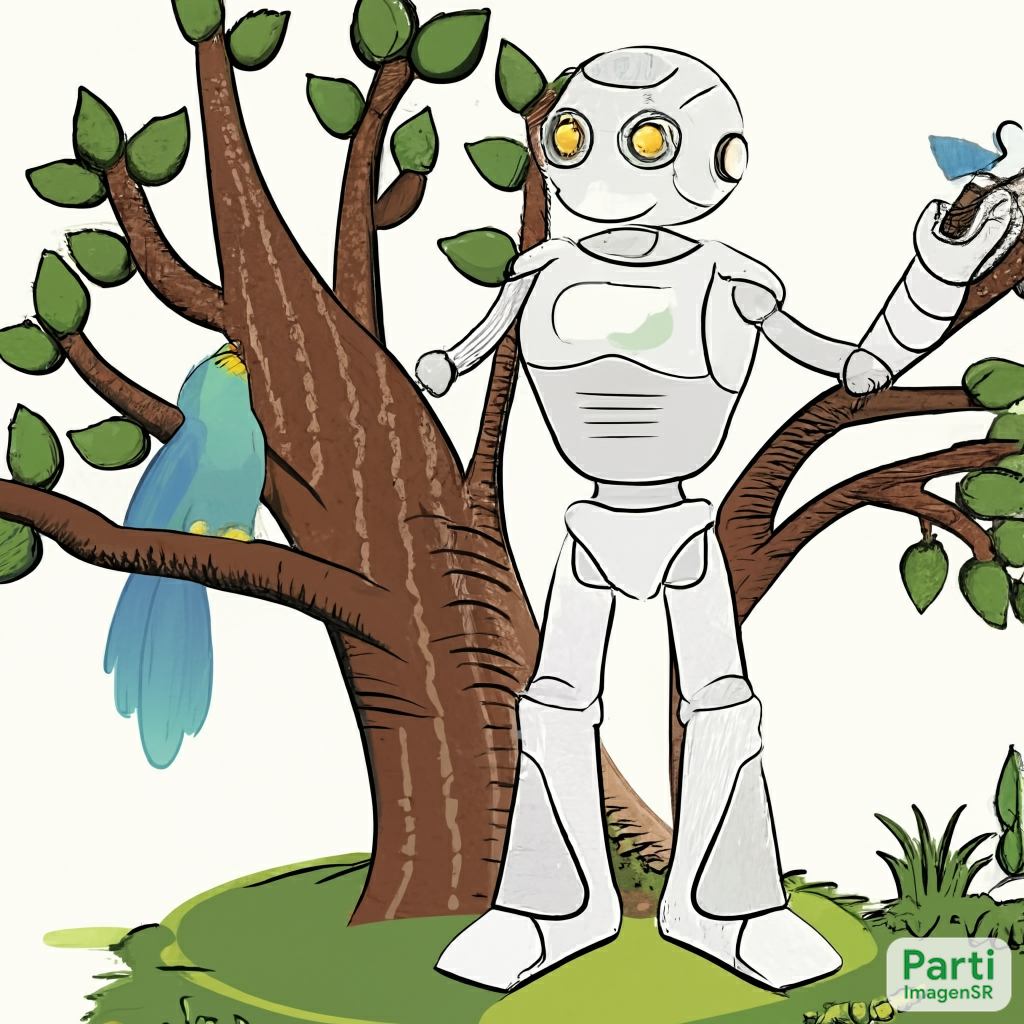}}}
     \end{subfigure}
     \hfill
    \caption{Generated depictions of the scene \textinput{A robot and its pet in a tree.} 
    Many elements are underspecified in the text, e.g., pet type, perspective, and visual style.
    }
    \label{fig:intro}
    \vspace{-0.4cm}
\end{figure}

With such developments in multimodal modeling and further aspirations towards contextualized language understanding, it is import to better understand both task validity and construct validity in text-to-image systems \cite{raji2021ai}. Ethical questions concerning bias, safety and misinformation are increasingly recognized \cite{saharia2022photorealistic,cho2022dall}; nevertheless, understanding which system behaviors are desirable requires a vocabulary and framework for understanding the diverse and quickly expanding capabilities of these systems.
This position paper addresses these issues by focusing on classic problems (in both linguistic theory and NLP) of ambiguity and underspecification \cite[e.g.,][]{poesio1994ambiguity, copestake2005minimal, frisson2009semantic}.
Little previous work has looked into how underspecification impacts multimodal systems, or what challenges and risks they pose. 

This position paper presents a model of task formulation in text-to-image tasks by considering the relationships between images and texts.
We use this foundation to identify challenges and risks when generating images of scenes from text descriptions, and discuss possible mitigations and strategies for addressing them. 

\section{Background}

\begin{figure*}
\small
    \centering
    \begin{tabular}{
|ll|}

\hline
\multicolumn{2}{|l|}{\cellcolor{gray!20}{Families of multimodal (text and image) tasks}} \\
\hline
  \begin{tabular}[t]{| p{6.5cm} |}
  \hline
 \cellcolor{gray!20}{Image-to-text tasks (\includegraphics[width=10pt]{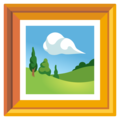}$\rightarrow$\includegraphics[width=10pt]{images/Text})} \\\hline
 Generating descriptions of scenes\\
 Optical character recognition\\
 Search index term generation\\
 $\cdot\cdot\cdot$\\
 \hline
  \end{tabular} 
 &
  \begin{tabular}[t]{| p{6.5cm} |}
  \hline
 \cellcolor{gray!20}{Text-to-image tasks (\includegraphics[width=10pt]{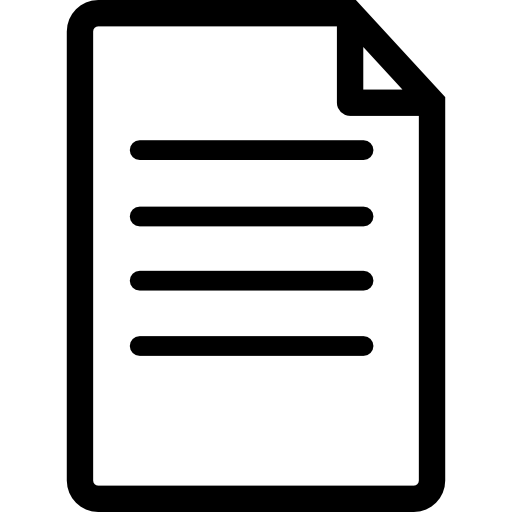}$\rightarrow$\includegraphics[width=10pt]{images/framed-picture.png})} \\\hline
 Generating depictions of scenes\\
 Story illustration\\
 Art generation\\
 $\cdot\cdot\cdot$\\
 \hline
  \end{tabular} \\
  \begin{tabular}[t]{| p{6.5cm} |}
  \hline
   \cellcolor{gray!20}{Image+text-to-text tasks
   (\includegraphics[width=10pt]{images/framed-picture.png}$+$\includegraphics[width=10pt]{images/Text.png}$\rightarrow$\includegraphics[width=10pt]{images/Text.png})} \\\hline
 Visual question answering\\
 $\cdot\cdot\cdot$\\
 \hline
  \end{tabular} 
  &
  \begin{tabular}[t]{| p{6.5cm} |}
\hline
 \cellcolor{gray!20}{Image+text-to-image tasks
    (\includegraphics[width=10pt]{images/framed-picture.png}$+$\includegraphics[width=10pt]{images/Text.png}$\rightarrow$\includegraphics[width=10pt]{images/framed-picture.png})} \\\hline
 Image editing using verbal prompts\\
 $\cdot\cdot\cdot$\\
 \hline
  \end{tabular} \\[9ex]
\hline
\end{tabular}
    \caption{Sketch of a taxonomy of text+image tasks. The taxonomy has gaps which suggest novel tasks, e.g., ``optical character generation'' (generating images of texts), or querying text collections using images.}
    \label{fig:tasks_taxonomy}
        \vspace{-0.4cm}

\end{figure*}

\subsection{Image meanings}
\label{sec:meaning}

Like texts, images are used in communicative contexts to convey concepts. Images often convey meaning via resemblance, whereas the correspondence between language and meaning is largely conventional (``icons'' vs ``symbols'' in the vocabulary of semiotics \cite[e.g.][]{saussure1916course, pierce1958collected, jappy2013introduction, chandler2007semiotics}). For example, both the English word ``cat'' or images of a cat---including photographs, sketches, etc---can signify the concept of a cat.
Furthermore they each can be used in contexts to represent either the general concept of cats, or a specific instance of a cat. That is, images  can have both i) concepts/senses, as well as ii) objects/referents in the world.
As such, both images and text can direct the mind of the viewer/reader towards objects and affairs in the world (also known as ``intentionality'' in the philosophy of language \citep[e.g.,][]{searle1995construction}), albeit in different ways.
Despite the adage that a picture is worth a thousand words, even relatively simple diagrams may not be reducible to textual descriptions \cite{griesemer1991must}.

Like texts, images can also indirectly convey meaning about the agent who produced the image, or about the technology used to create or transmit it \citep[cf. the model of communication of][]{jakobson1960closing}. 
Also like language, the meanings of images can be at least partly conventional and cultural, e.g., logos, iconography, tattoos, crests, hand gestures, etc.\ can each convey meaning despite having no visual resemblance to the concept or thing being denoted. \citet{shatford1986analyzing} describes this in terms of images being \emph{Of} one thing yet potentially \emph{About} another thing.
Such ``aboutness'' is not limited to iconography, for photographic imagery can convey cultural meanings too---\citet{barthes1977image} uses the example of a photograph of a red chequered tablecloth and fresh produce conveying the idea of Italianicity. 

\subsection{Text-image relationships}
\label{sec:relationships}

A variety of relationships between text and image are possible, and have been widely discussed in creative and cultural fields  \cite[e.g.,][]{barthes1977image,berger2008ways}. The Cooper Hewitt Design Museum has, for example, published extensive guidelines on accessible image descriptions.\footnote{\url{https://www.cooperhewitt.org/cooper-hewitt-guidelines-for-image-description/}} These make a fundamental distinction between image \emph{descriptions}, which provide visual information about what is depicted in the image, and \emph{captions}, which explain the image or provide additional information. For example, the following texts could apply to the same image, while serving these different purposes:
\begin{itemize}[noitemsep,leftmargin=*,parsep=0pt,partopsep=0pt]
    \item \textbf{description}: \textinput{Portrait of former First Lady Michelle Obama seated looking directly at us.}
    \item \textbf{caption}: \textinput{Michelle LaVaughn Robinson Obama, born 1964, Chicago, Illinois.} 
\end{itemize}
\noindent
This distinction is closely related to that between \emph{conceptual descriptions} and \emph{non-visual descriptions} made by \citet{hodosh2013framing}, building on prior work on image indexing \cite{jaimes2000conceptual}. Hodosh \etal subdivide conceptual descriptions into \emph{concrete} or \emph{abstract} according to whether they describe the scene and its entities or the overall mood, and also further differentiate a category of \emph{perceptual descriptions} which concern the visual properties of the image itself such as color and shape. \citet[Chapter 2]{van2019pragmatic} has a more detailed review of these distinctions.

As images have meanings (see \cref{sec:meaning}), describing an image often involves a degree of interpretation \cite{van2020use}. Although often presented as neutral labels, captions on photographs commonly tell us how visual elements ``ought to be read'' \cite[p.\ 229]{hall2019determinations}.
Literary theorist Barthes distinguishes two relationships between texts and images: \emph{anchorage} and \emph{relay}. With anchorage, the text guides the viewer towards certain interpretations of the image, whereas for relay, the text and image complement each other
\cite[pp.\ 38--41]{barthes1977image}. 
McCloud's theory of comics elaborates on this to distinguish four flavours of word-image combinations \cite{mccloud1993understanding}: 
(1) the image supplements the text,
(2) the text supplements the image,
(3) the text and image contribute the same information,
(4) the text and image operate in parallel without their meanings intersecting.
Since language is interpreted contextually, these image-accompanying texts might depend on the multimodal discourse context, the writer, and the intended audience. 
The strong dependence on the writer, in particular, highlights the socially and culturally subjective nature of image descriptions \cite{van2017cross, bhargava2019exposing}. This subjectivity can result in speculation (or abductive inference), for example when people describing images fill in missing details \cite{van2020use}, in human reporting biases regarding what is considered noteworthy or unexpected \cite{van2016pragmatic, misra2016seeing}, in social and cultural stereotyping \cite{van2016stereotyping, zhao2017men, otterbacher2019we}, and in derogatory and offensive image associations  \cite{birhane2021multimodal,crawford2019excavating}.

Despite the frequently stated motivation of ML-based multimodal image+text technologies as assisting the visually impaired, the distinction between captions and descriptions---relevant to accessibility---is mostly ignored in the text-to-image literature \cite{van2019pragmatic,van2020use}. 
It is common for systems that generate image descriptions to be described as ``image-captioning'' \cite[e.g.,][]{nie2020pragmatic, agrawal2019nocaps, srinivasan2021wit, lin2014microsoft, sharma2018conceptual}, without making a distinction between captions and descriptions. An exception is a recent paper explicitly aimed at addressing image accessibility \cite{kreiss2021concadia}. Other NLP work uses ``caption'' to denote characterizations of image content, using ``depiction'' for more general relations between texts and images \cite{alikhani2019caption}.

Within multimodal NLP, building on annotation efforts, Alikhani \etal have distinguished five types of coherence relationships in aligned images and texts (of which multiple can hold concurrently) \cite{alikhani2020clue, alikhani2019cite}:
(1) the text presents what is depicted in the image,
(2) the text describes the speaker’s reaction to the image,
(3) the text describes a bigger event of which the image captures only a moment,
(4) the text describes background info or other circumstances relevant to the image, and
(5) the text concerns the production and presentation of the image itself.

Finally, we also note the case where the image is of (or contains) text itself. Not only is this relevant to OCR tasks, but also to visual analysis of web pages \cite[e.g.,][]{mei2016pagesense}, memes \cite[e.g.,][]{kiela2020hateful}, advertising imagery \cite[e.g.,][]{lim2017multimodal}, as well as a challenging aspect of image generation when the image is desired to have embedded text (for example on a book cover).
(Prior to movable type printing, the distinction between texts and images-of-texts was likely less culturally important  \cite{ong2013orality, sproat2010language}.)

\subsection{Text-to-image tasks}

Figure~\ref{fig:tasks_taxonomy} situates the family of text-to-image tasks within the greater family of multimodal (text and image) tasks. One of the important factors distinguishing different flavors of text-to-image tasks is the semantic and pragmatic relationship between the input text and the output image. 
Although commonly used as if it describes a single task, we posit that
``text-to-image'' describes a family of tasks, since it only denotes a structural relationship: a text goes in and an image comes out. Although some relationship between input and output is perhaps implied, it is just as implicit as if one were to speak of a ``text-to-text'' task without mentioning whether the task involves translation, paraphrase, summarization, etc. It is important to emphasise that tasks and models are typically not in a 1:1 relationship: even without multi-head architectures, a model may be used for many tasks \cite[e.g.,][]{raffel2020exploring,chen2022pali}, while many (single-task) NLP architectures employ multiple models in sequence.  As \citet{van2020use} argues, the dataset annotations which often act as extensional definitions of the task of interest \cite{schlangen2021targeting} are often produced via underspecified crowdsourcing tasks that do not pay full attention to the rich space of possible text-image relationships described above.
Similarly, text-image pairs repurposed from the web often have poorly specified relationships:
although the Web Content Accessibility Guidelines recommend that ``alt'' tags ``convey the same function or purpose as the image'' \cite{chisholm2001web} (for a survey of guidelines, see \citet{craven2006some}),
real-world usage may deviate considerably (see, e.g., \cite{petrie2005describing} and the discussion in \cite{muehlbradt2022s}).

Recent literature on text-to-image modeling has been characterized by simplified task formulations. For example, despite the impressive outputs of recent models---e.g., un\textsc{clip} (a.k.a., \textsc{dall-e 2}) \cite{ramesh2022hierarchical}, Imagen \cite{saharia2022photorealistic}, and Parti \cite{yu2022scaling}---the papers
introducing these models rely on the broadest task formulation, wherein the model takes a textual prompt of any kind and produces an image of any kind. While they discuss terms such as \textit{diversity}, \textit{caption similarity}, \textit{high fidelity}, and \textit{high quality} to discuss properties of model outputs, these are not precisely defined, nor are they fully operationalized in current evaluation metrics.
Similarly, the \textsc{xmc-gan} paper asserts that systems should produce ``coherent, clear, photo-realistic scenes'' yet the authors fail to either justify or clarify these objectives \cite{Zhang_2021_CVPR}. In fact, this objective seems to be at least partly a by-product of the fact that the model training and evaluation was on photographs from the \textsc{ms-coco} dataset.  Setting photo-realistic imagery as the ideal raises questions about both justification ({why not other styles of images?}) and correspondence (e.g., {how does photography construct relationships between images and reality?}). 

\section{Task Formulation}
\label{sec:formulation}

Underspecification in task formulation is a major challenge for machine learning and artificial intelligence disciplines as a whole \cite{damour2020underspecification, raji2021ai}. 
Clarity around task formulation helps system designers navigate ambiguous inputs; for example, given a prompt such as \textinput{a painting of a horse}, should the system create an image whose style resembles a painting, or an image of a scene containing a painting, including the frame and other plausible contextual details?  This paper postulates that accounts of \emph{image meaning} and \emph{text-image relationships} are of central relevance to formulating task definitions in text-to-image systems generally. Such accounts are thus important for characterizing underspecification in such systems.

\begin{figure}
    \centering
    \includegraphics[width=0.45\textwidth]{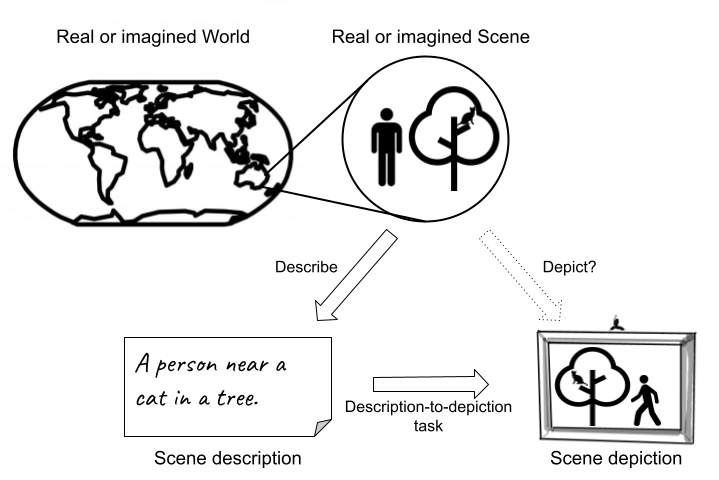}
    \caption{Scene depictions and descriptions are communicative acts conveying information (or misinformation) about a real or imagined scene in the world.
    }
    \label{fig:image text world}
        \vspace{-0.4cm}
\end{figure}

We take the notion of \emph{world} to be important too, for two reasons. Like texts, images can reference objects in the world, and in doing so are human-mediated representations of the observable world that involve selection and filtering processes. Also, the notion of possible worlds has played an important role in theories of semantics (e.g., \citet{kratzer1998semantics}).
Therefore, two questions that we believe should be central to an account of underspecification in text-to-image tasks are: 

\begin{enumerate}[noitemsep,leftmargin=*,parsep=0pt,partopsep=0pt]
    \item What are the two-way relationships between images-text pairs and (real or imagined) worlds? 
    \item What is the three-way relationship between the images, texts and the world? 
\end{enumerate}

\noindent
We do not attempt here to unify or rebut the many theories of image meanings and text--image relationships, but instead highlight what we see as essential considerations for scene depiction tasks:
\begin{enumerate}[noitemsep,leftmargin=*,parsep=0pt,partopsep=0pt]
    \item We use \emph{scene} to mean a small fragment of a (real or imagined) world. A scene can be \emph{described} in texts, and can also be \emph{depicted} in images.
    Both descriptions and depictions can thus convey information about a scene.\footnote{We use ``depiction'' in the sense of ``to show visually'', rather than the definition by \citet{alikhani2019caption}.}
    \item The production and sharing of descriptions and depictions both constitute \emph{communicative acts}. These acts are interpreted within social contexts, and can have locutionary (what is said/shown) as well as perlocutionary dimensions (effects on the viewer/reader such as scaring, offending or prompting action) and illocutionary dimensions such as connotations. 
    \item Descriptions and depictions necessarily convey \emph{incomplete} information about all but the most trivial scene.
    The two modalities necessarily \emph{underspecify} different types of information, both due to intra-modal constraints and assumptions of extra-modal contextual information.
\end{enumerate}

We propose two components, \textit{coherence} and \textit{style}, for the formulation of the family of text-to-image tasks. We argue in the following section that both are relevant to underspecification.

\begin{itemize}[noitemsep,leftmargin=*,parsep=0pt,partopsep=0pt]
    \item \textbf{Coherence}: Any valid semantic and/or pragmatic relationship between a static image-text pair, e.g., those listed in  \cref{sec:relationships}, is a potentially valid semantic relationship for a given flavor of text-to-image task. For example, one can meaningfully speak of a description-to-depiction task or an event-to-image-moment task.
    \item \textbf{Style}: Valid text-to-image tasks can encompass a multitude of visual styles. That is, text-to-image is not constrained to photo-realism but rather can involve styles resembling {cartoons}, {paintings}, {woodcut prints}, etc, and even to specific genres such as {manga}, {impressionist}, or {ukiyo-e}.
\end{itemize}

Given this conceptual framework, one natural challenge that presents itself is that visual and linguistic information often serve to complement each other in multimodal texts. Indeed this can be utilized for skilled effect leading to greater engagement with readers/viewers by requiring that they mentally fill in the missing information \cite{mccloud1993understanding, iyyer2017amazing}. 

\section{Challenges in Description to Depiction}

\begin{figure*}
    \centering
         \begin{subfigure}[t]{0.24\textwidth}
         \centering
         \includegraphics[width=0.8\textwidth]{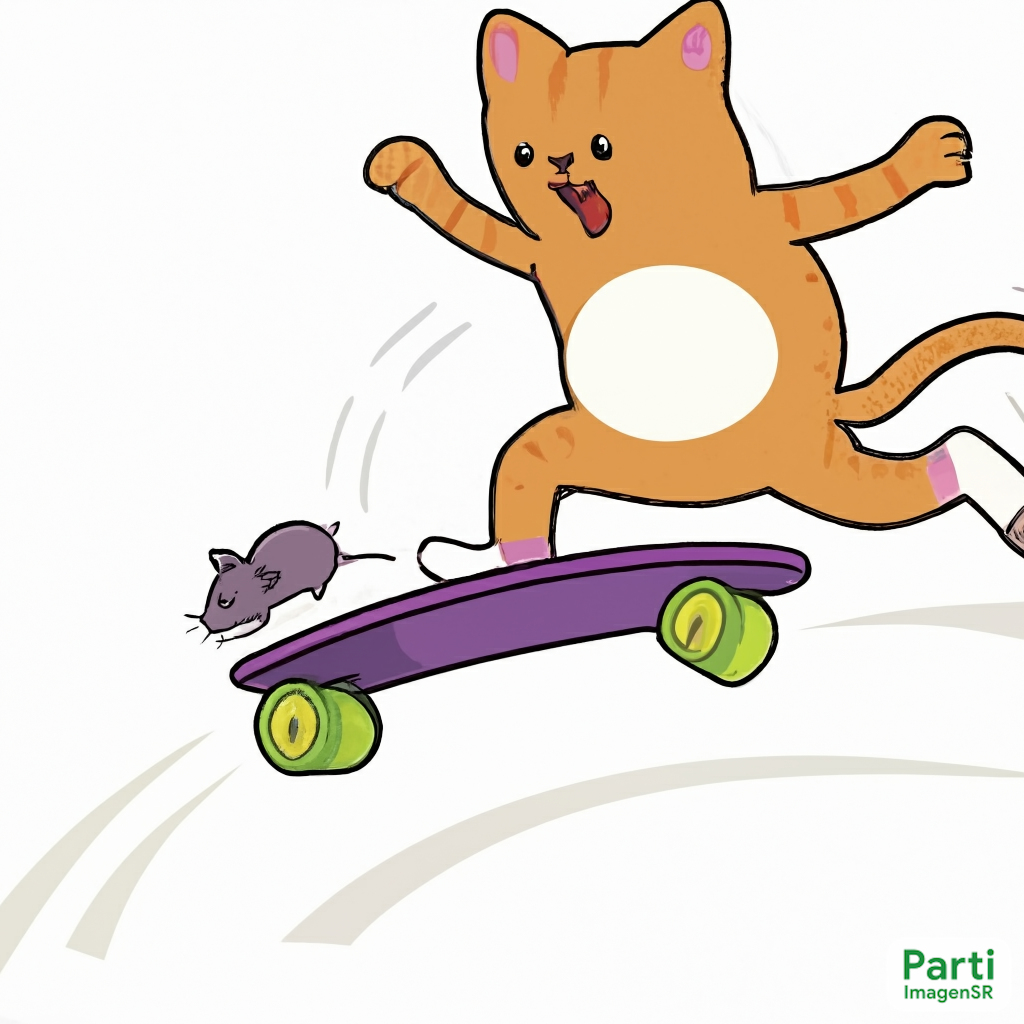}
         \includegraphics[width=0.8\textwidth]{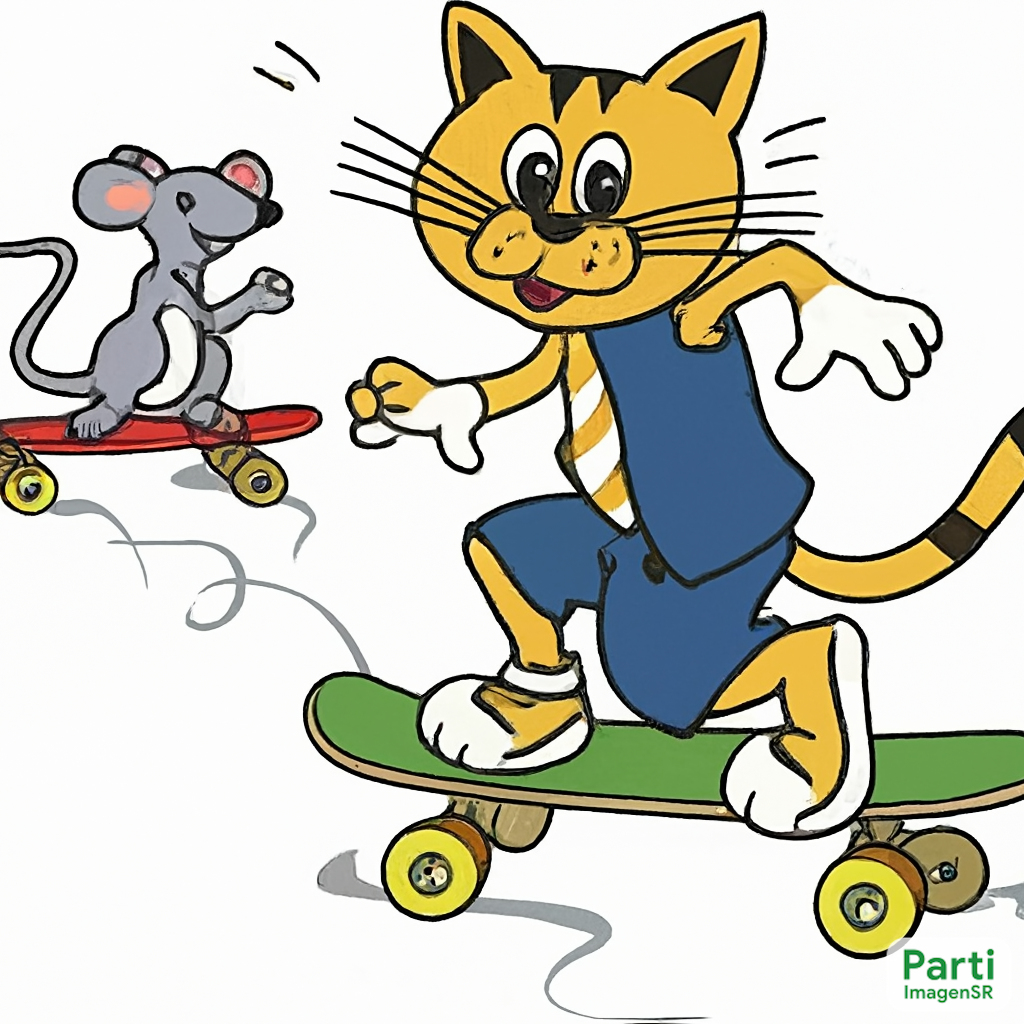}
         \caption{Outputs for \textinput{A cat chasing a mouse on a skateboard.}
         The number of boards and which animal is on any given board is ambiguous.
         }
    \label{fig:examples+underspecified_a}
     \end{subfigure}
     \hfill
         \begin{subfigure}[t]{0.24\textwidth}
         \centering
         \includegraphics[width=0.8\textwidth]{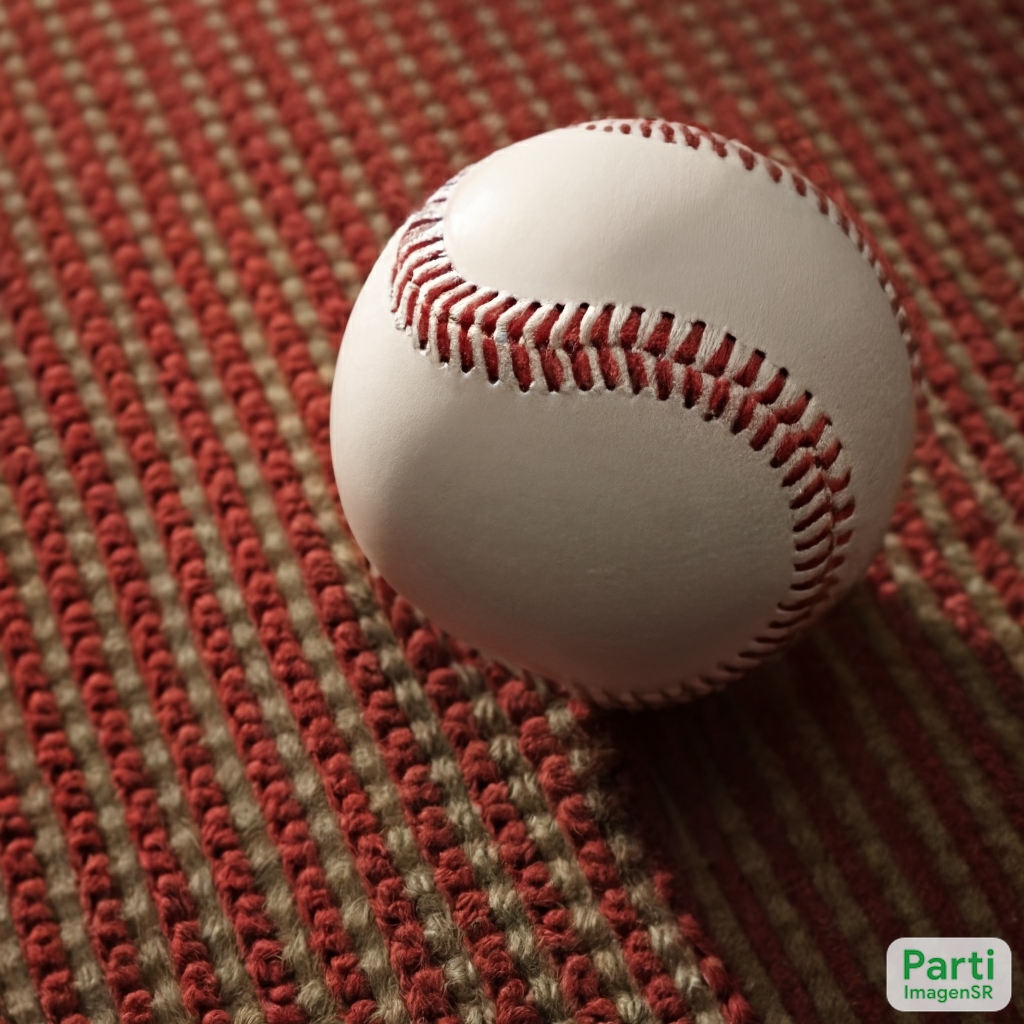}
         \includegraphics[width=0.8\textwidth]{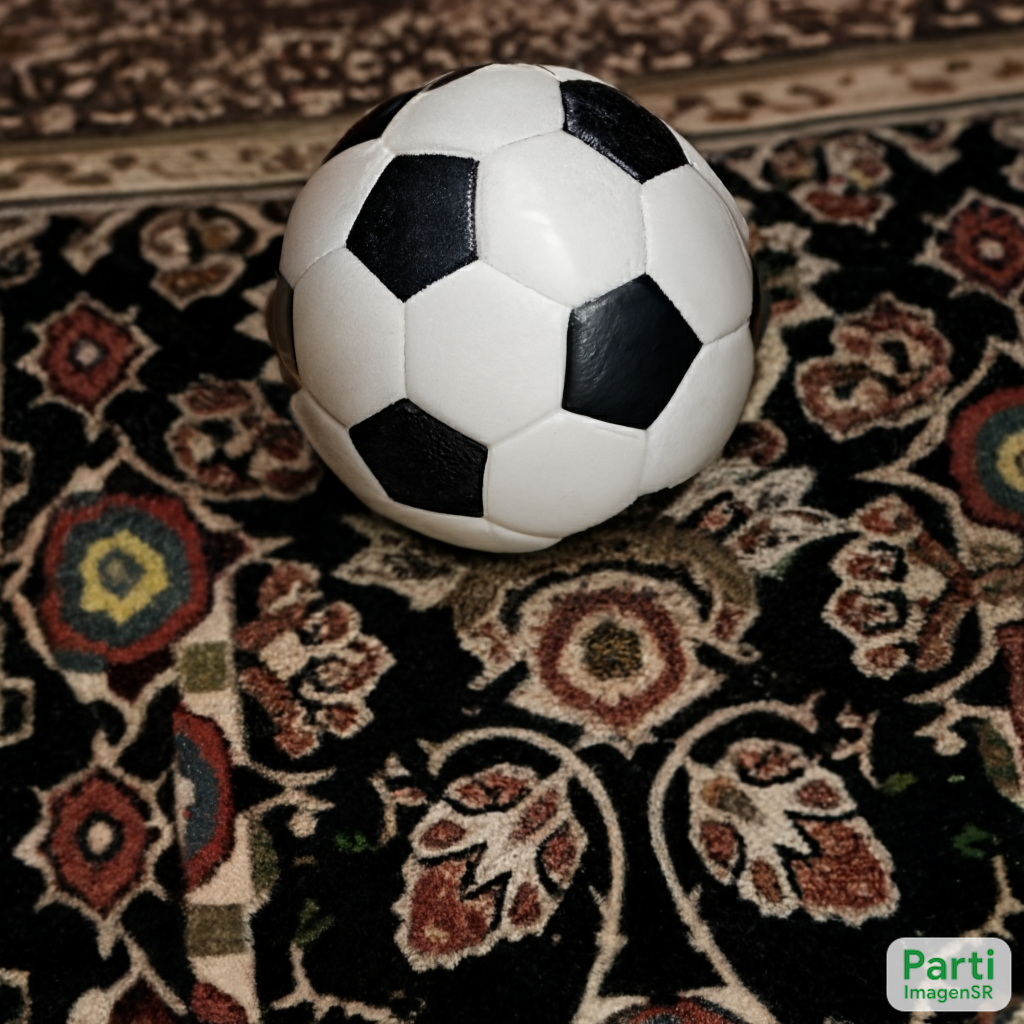}
         \caption{Outputs for \textinput{A ball on a rug.} The types and visual details of balls and rugs are unspecified.}
    \label{fig:examples+underspecified_b}
     \end{subfigure}
     \hfill
         \begin{subfigure}[t]{0.24\textwidth}
         \centering
         \includegraphics[width=0.8\textwidth]{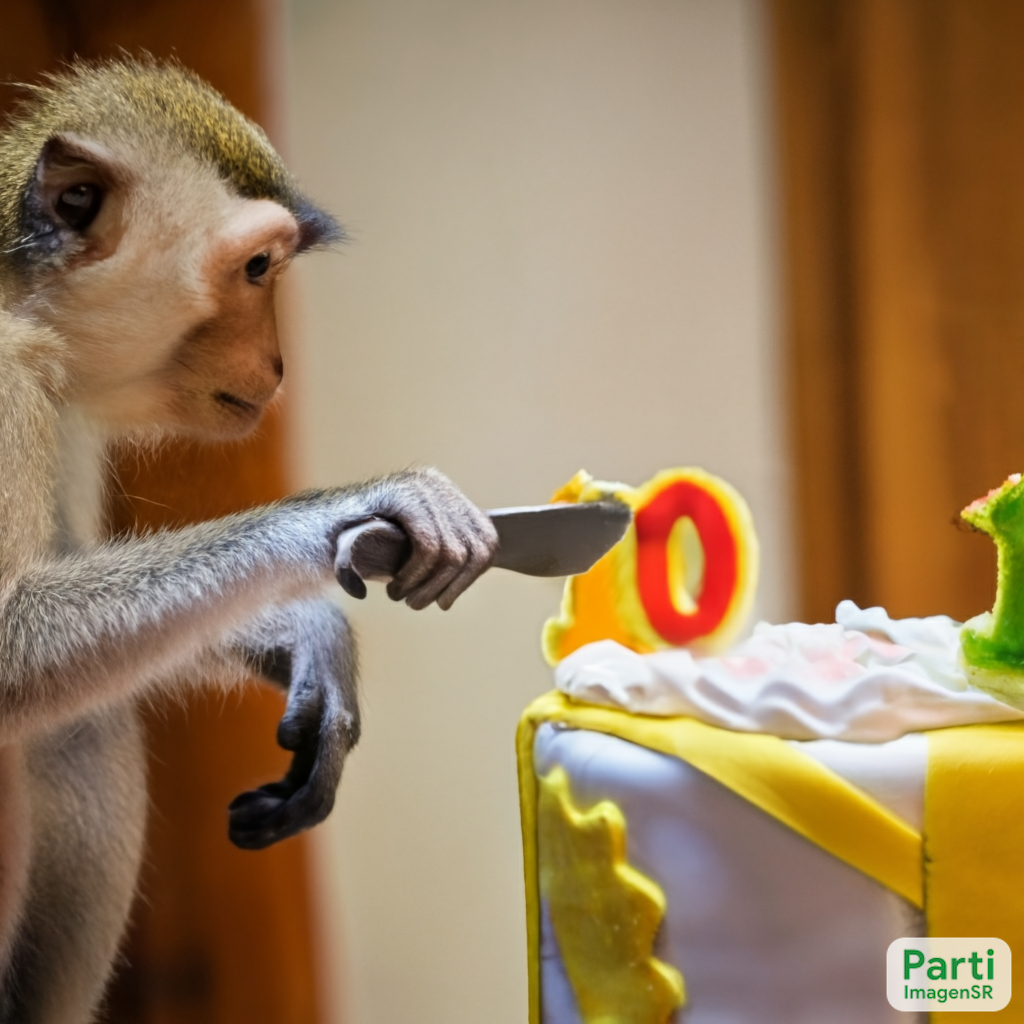}
         \includegraphics[width=0.8\textwidth]{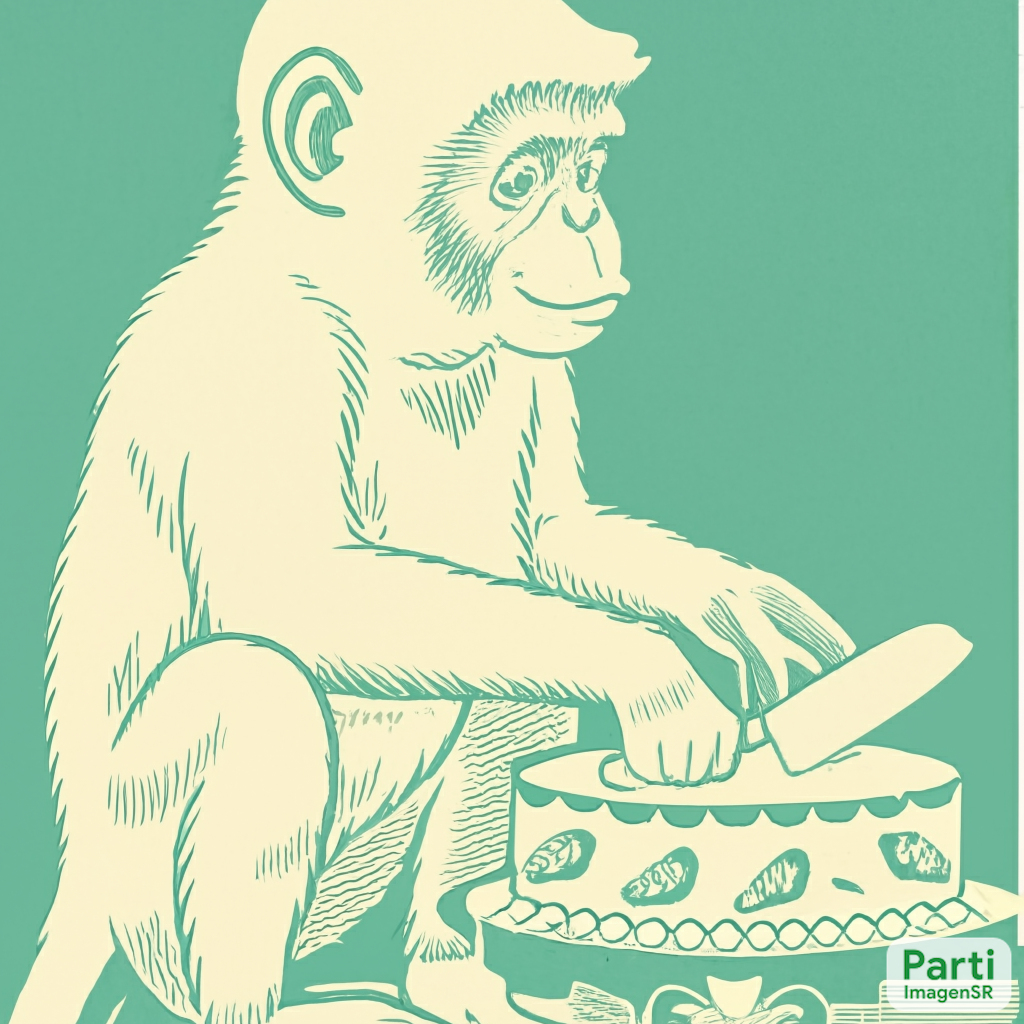}
         \caption{Outputs for \textinput{A monkey cutting a cake.} The cutting instrument is unspecified, as is the style.}
             \label{fig:examples+underspecified_c}

         \end{subfigure}
         \hfill
         \begin{subfigure}[t]{0.24\textwidth}
         \centering
         \includegraphics[width=0.8\textwidth]{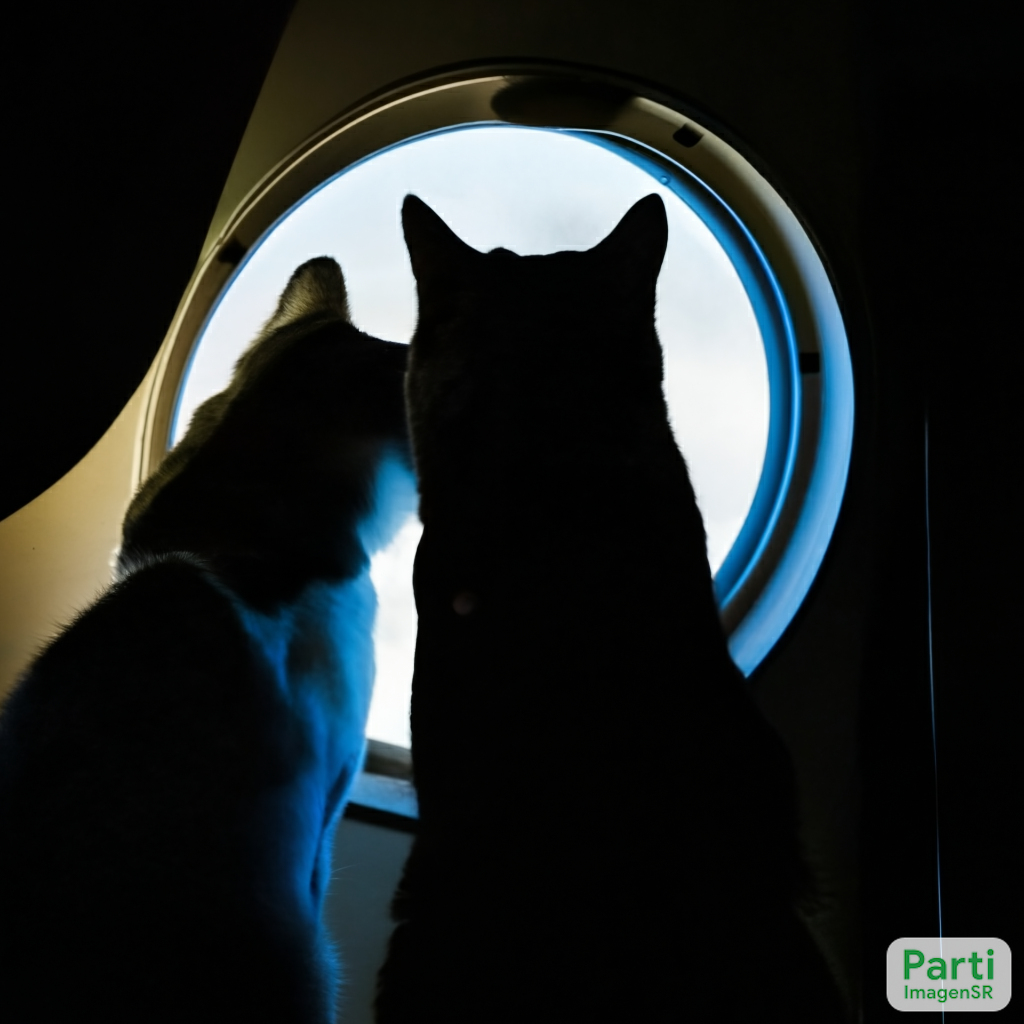}
         \includegraphics[width=0.8\textwidth]{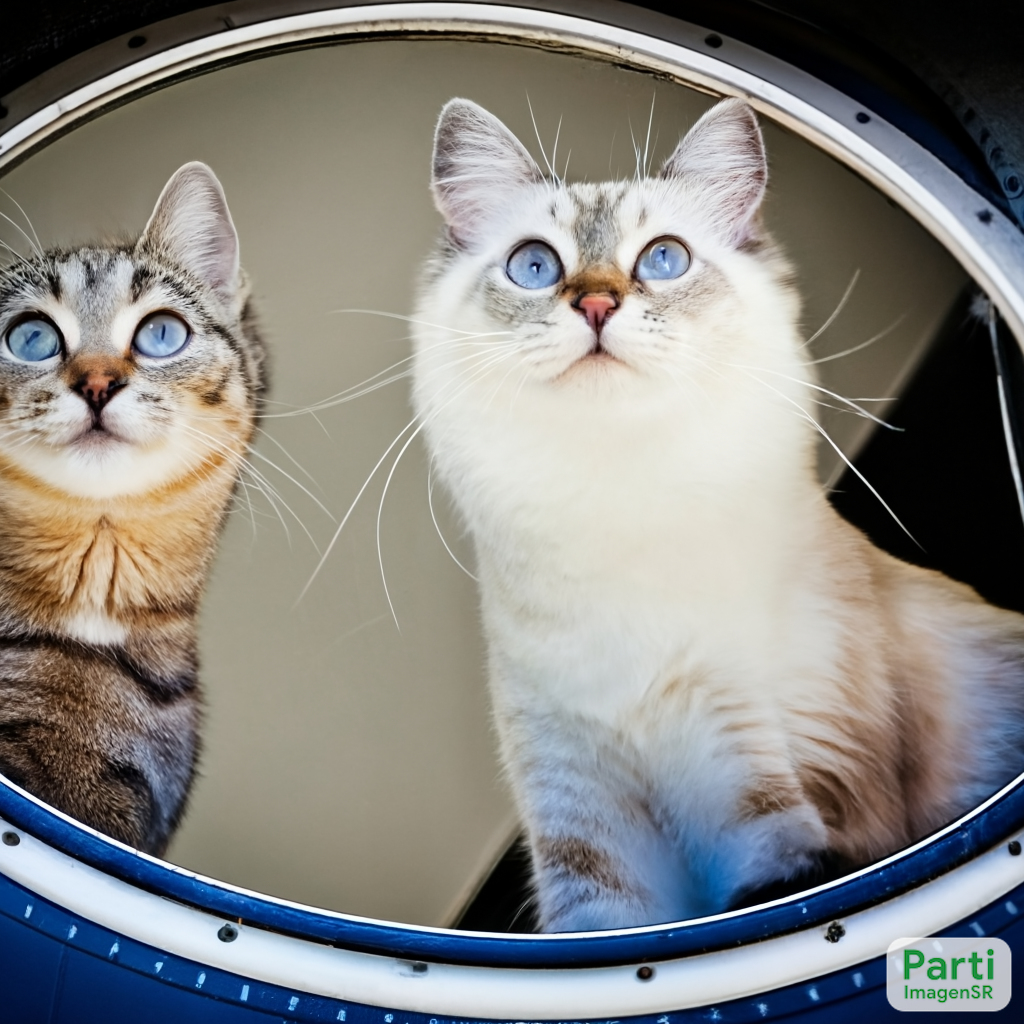}
         \caption{Outputs for \textinput{Two cats looking out of a space shuttle window. DSLR photograph.} Perspective is unspecified.}
         \label{fig:examples+underspecified_d}
         \end{subfigure}
     \hfill
     \caption{Example treatments of underspecified inputs.
      These examples and those elsewhere in this paper were generated using Parti \cite{yu2022scaling} followed by the super-resolution third stage of Imagen \cite{saharia2022photorealistic}.
     }
    \label{fig:examples+underspecified}
    \vspace{-0.4cm}
\end{figure*}

Having laid out the relevant considerations of meaning and reference in text-to-image systems in \cref{sec:formulation}, we now focus specifically on systems that produce an image depiction of a scene from a description of that scene. We distinguish challenges from three sources: {linguistic ambiguity in descriptions}, {underspecification in descriptions}, and {underspecification of desired depictions}.
Our use of the term \emph{underspecification} here reflects how it has been used in NLP literature, referring both to ambiguity in the objects of study (e.g., linguistic forms  \cite[p.\ 29]{bender2019linguistic}), as well as to properties of the technical apparatus used to model meaning (e.g., \citealt[p.\ 30]{bender2019linguistic}).

\subsection{Linguistic Ambiguity in Descriptions} Many if not all forms of linguistic ambiguities are likely to occur in scene descriptions. However we call out a few of notable importance. 
\begin{itemize}[noitemsep,leftmargin=*,parsep=0pt,partopsep=0pt]
    \item \textit{Syntactic ambiguities} including locative PP attachment can present ambiguities concerning spatial relationships. For instance, in the input \textinput{A cat chasing a mouse on as skateboard}, is the cat or the mouse---or both---on the skateboard? See Figure~\ref{fig:examples+underspecified_a}.
    \item \textit{Word sense ambiguities} (including metonymy) and ontological vagueness present challenges as to how objects should be depicted; e.g., for \textinput{The man picked up the bat}, is the bat a flying mammal or a sports implement? Visualizing ambiguous words is also a challenge for verbs:  \textinput{riding a bus} and \textinput{riding a horse} are very different actions (consider that ``riding a bus in the way one would normally ride a horse'' is easier to imagine than the converse) \cite{gella2017disambiguating}. 
    \item \textit{Anaphoric ambiguities} including pronouns can also cause challenges, e.g., what is the toy beside in \textinput{a book on a chair and a toy beside it}? \item \textit{Quantifier scope ambiguities} also arise, e.g., how many books are there in \textinput{three people holding a large book}? 
\end{itemize}

\subsection{Underspecification in Descriptions}

Finite and reasonable-length linguistic descriptions of real-world or realistic scenes will by necessity omit a great deal of visual information.
Within NLP, underspecification in descriptions has perhaps been discussed most often in the context of generating referring expressions for objects (see \citet{krahmer2012computational} for a survey).
However, underspecification in input texts also causes major challenges in description to depiction tasks. 


\begin{itemize}[noitemsep,leftmargin=*,parsep=0pt,partopsep=0pt]

\item \textit{Unmarked defaults} can lead to potentially unbounded amounts of underspecified information (e.g., should people be depicted as clothed even if clothing is not mentioned, as is the social norm in images?) \cite{misra2016seeing}. Visual details such as lighting, color and texture may be omitted from texts: What does a carpet's surface look like? Where is the light source and do shadows need to be depicted?). See Figure~\ref{fig:examples+underspecified_b}.

\item \textit{Ontological vagueness} may also present challenges as to what types of objects should be depicted: for \textinput{a tall dark-skinned person with a toy}, what type of toy? 
See also Figure~\ref{fig:examples+underspecified_b}. Scalars typically often present underspecification (e.g., how tall is \textinput{tall person}?; how dark is \textinput{dark-skinned}?), and points of reference are often underspecified (cf. \textinput{tall} and \textinput{dark-skinned} in Japan vs South Africa). Ontological specificity in texts depends at least partly on which categories are considered to be basic \cite[e.g.][]{rosch1976basic, ordonez2015predicting}.

\item \textit{Geo-cultural context} of input descriptions is often left unspecified. For instance, in \textinput{a woman eating breakfast beside her pet}, the types of things that count as breakfast and pets are culturally subjective. In many cases, object forms are institutionally regulated, e.g., for \textinput{a man counting money in a car}, the physical appearance of money and license plates, and the positioning of the steering wheel (left vs. right), are institutionally regulated and only implicit in the text.

\item \textit{Implied objects} that are part of many events or states are often not specified in corresponding descriptions. For example \textinput{a monkey cutting a cake} implies a cutting instrument (see Figure~\ref{fig:examples+underspecified_c}); \textinput{a wedding} has many implied objects, but at a minimum seems to imply two people.

\end{itemize}

While description to depiction models often generate images that fills in such implied details or objects, such extrapolations run the risk of perpetuating social stereotypes (\cref{sec:risks}).

\begin{figure*}
    \centering
         \begin{subfigure}[t]{0.24\textwidth}
         \centering
         \includegraphics[width=0.8\textwidth]{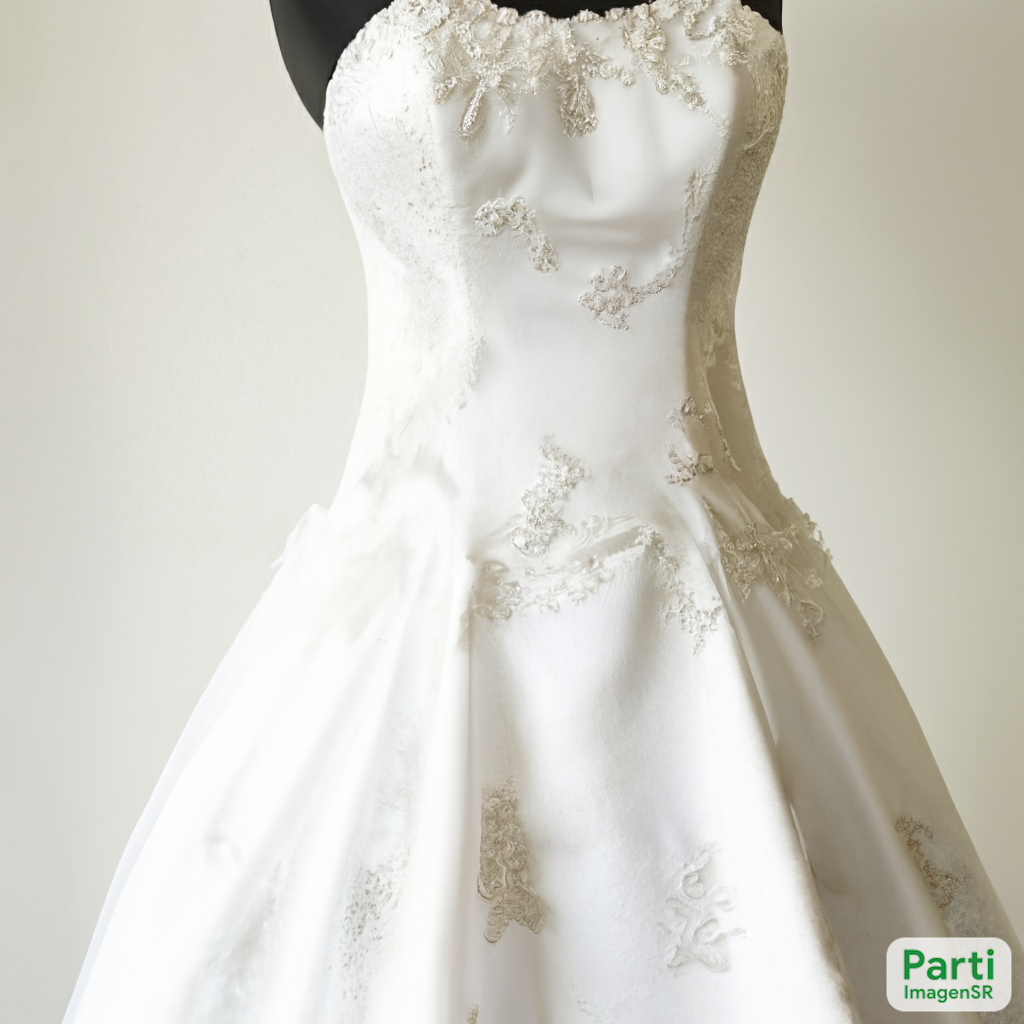}
         \caption{Outputs for \textinput{Wedding attire displayed on a mannequin} may show gender and Western cultural biases.}
    \label{fig:examples_risks_a}
     \end{subfigure}
     \hfill
         \begin{subfigure}[t]{0.24\textwidth}
         \centering
         \includegraphics[width=0.8\textwidth]{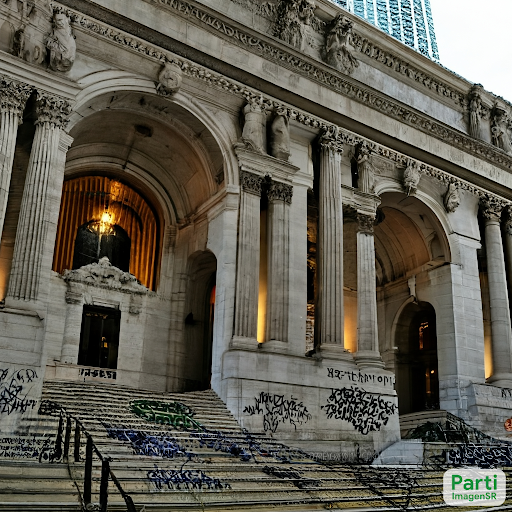}
         \caption{Outputs for \textinput{Graffiti on the New York Public library. DSLR photo.} might cause offence to bibliophiles.}
    \label{fig:examples_risks_b}
    \end{subfigure}
     \hfill
         \begin{subfigure}[t]{0.24\textwidth}
         \centering
         \includegraphics[width=0.8\textwidth]{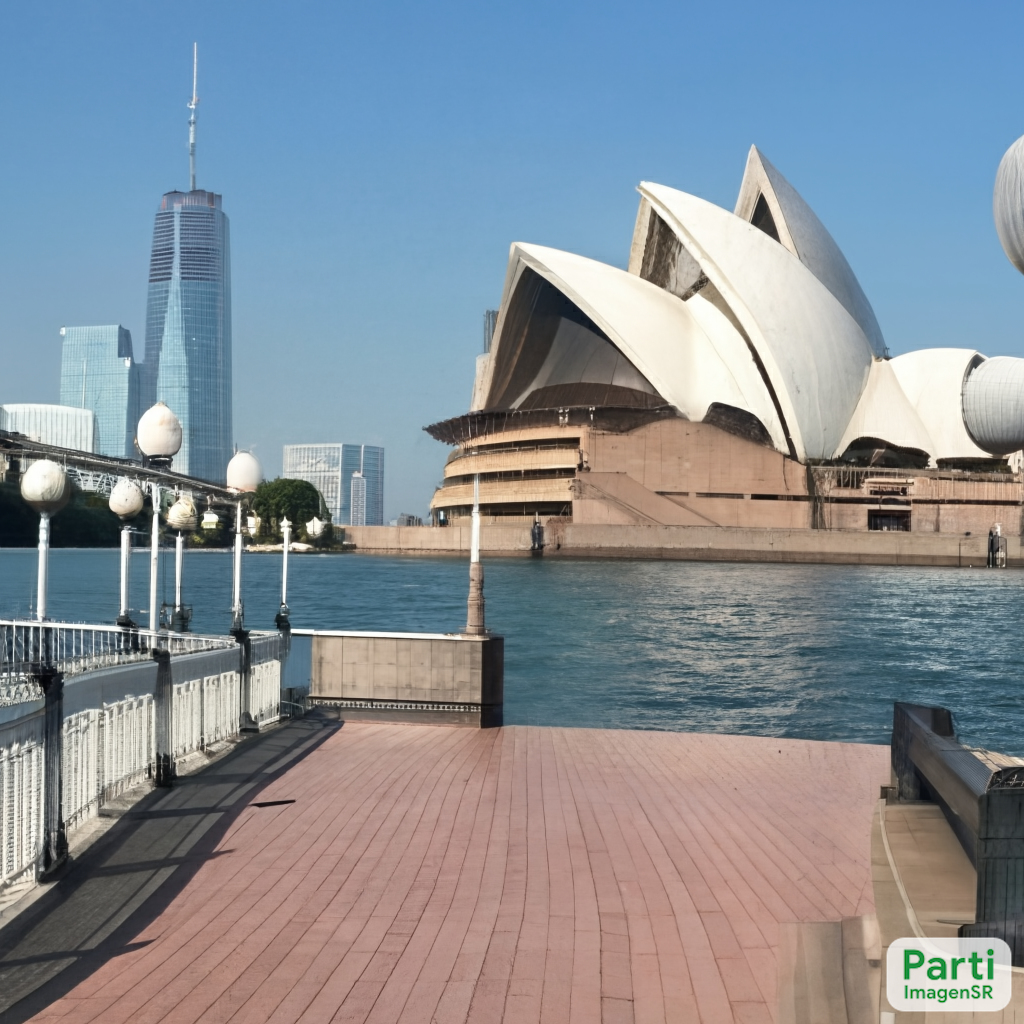}
         \caption{Outputs for \textinput{A photo of a famous city with opera house} may spread misinformation.}
    \label{fig:examples_risks_c}
     \end{subfigure}
     \hfill
         \begin{subfigure}[t]{0.24\textwidth}
         \centering
         \includegraphics[width=0.8\textwidth]{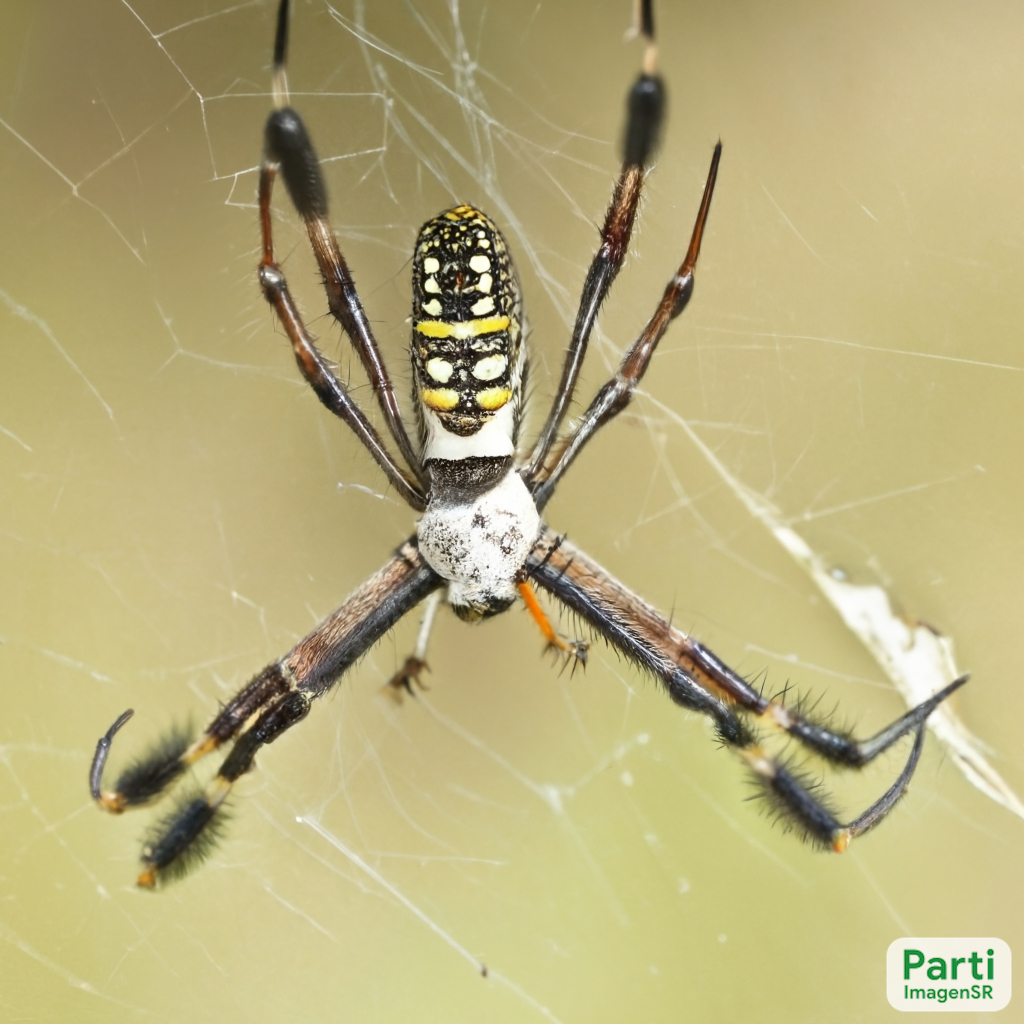}
         \caption{Outputs for \textinput{A photo of a non-venomous Australian spider} may have safety risks for animal lovers.}
    \label{fig:examples_risks_d}
     \end{subfigure}
     \hfill
     \caption{Example of risks in scene description-to-depiction.}
          \vspace{-0.4cm}

    \label{fig:examples_risks}
\end{figure*}

\subsection{Underspecification of Desired Depictions}

The underspecification challenges in the linguistic inputs to text-to-image systems are complemented by a different set of challenges in the output generation concerning precise visual details. 

\begin{itemize}[noitemsep,leftmargin=*,parsep=0pt,partopsep=0pt]

\item \textit{Style}. Text inputs often do not specify a desired visual style of depiction, e.g., {photo-realism}, {cartoons}, {paintings}, {woodcut prints}, etc., or genres such as {manga}, {impressionist}, and {ukiyo-e}. While this is a question relevant also for task formulation (see \cref{sec:formulation}), this ambiguity need to be resolved for text-to-image systems capable of generating multiple styles of images. It is also possible to imagine and create new styles using these tools. This is a fascinating use case, but it also raises questions about how to evaluate whether a model has succeeded---for example, when mashing together multiple style specifications, e.g. \textinput{The New York City skyline in ukiyo-e style by van Gogh.}

\item \textit{Technical}. Goals of photo(graphic)-realism  raise questions about what sort of photographic technologies are implied, including implicit lens, implicit depth of focus and implicit exposure time, each of which produce different visual artefacts.

\item \textit{Perspective}. Many image styles, including but not limited to photographic ones, have an implied perspective, and an implied frame or shot \cite[p.\ 89]{chandler2007semiotics}, including not just an implied eye but also an implied angle or tilt. The choice of perspective can have socio-cultural connotations. A perspective closer to the ground may represent that of a child, and low viewing angles are used by filmmakers to make subjects appear powerful or convey vulnerability.\footnote{\url{https://www.nfi.edu/low-angle-shot/}} Such low-shots might also impact subject credibility \cite{mandell1973judging}.
Different social groups may have proclivities for different angles \cite{y2017physiognomy} or perspectives (e.g., \citealt{green2009between}, discussed in \citealt{cohn2013visual}).
\item \textit{Spatial orientations} with respect to the implied viewer (see Figure \ref{fig:examples+underspecified_d}) are not typically mentioned in the image descriptions upon which models are trained. For example, it is common in a portrait for the subject to be oriented so their face is visible, however such orientation towards the viewer is often not made explicit.

\end{itemize}

\noindent Finally, we note that linguistic ambiguities can interact with underspecified perspectives. An example provided by \citet{levelt1999producing} is the congruity of an image with the text ``a house with a tree to the left of it'' depends not just on the perspective taken in framing the image, but also whether ``to the left of'' is with respect to the viewer's orientation (facing the house) or to the house's orientation (e.g., facing the viewer, if the front of the house is depicted).

\section{Risks and Concerns}
\label{sec:risks}

Some datasets used for training multimodal systems have previously been shown to contain biases, stereotypes and pornography \cite{birhane2021multimodal,van2016stereotyping}.
We now discuss potential concerns in applications employing scene description-to-generation tasks, including how underspecification challenges can exacerbate them.

\textbf{Bias:} As in image-to-text \cite{bennett2021s}, there are risks of text-to-image amplifying societal biases including those concerning gender, race, and disability. Since English-language texts do not grammatically require specification of gender identities of people mentioned in a scene, there is a great potential for systems to reproduce existing societal biases. For example, the prompt \textinput{a boss addressing workers} might produce an image of a boss with masculine phenotypes. Similar outcomes are likely to be obtained with respect to other social roles, social groups and stereotypical phenotypes. Cultural biases are expected to be prevalent in any text-to-image systems, since what events and artefacts look like vary wildly around the world---e.g., weddings, bank notes, places of worship, breakfast dishes, etc.
When a prompt is ambiguous or underspecified, an ML model is likely to revert to correlations in its training data for deciding details about objects and their appearances. Thus underspecification leads to a greater risk of stereotyping biases, which can cause offense and representational harm especially to marginalized groups with a history of being stereotyped. See Figure~\ref{fig:examples_risks_a}.

\textbf{Harmful, taboo and offensive content:} Images depicting violent scenes may have a greater impact on the viewer than corresponding text descriptions. Similarly, pornographic images can be more shocking or culturally taboo than texts. Some societies, such as indigenous Australian ones, may have taboos on visual depictions  of the recently deceased
\cite[p.\ 20]{2018greater}. This exemplifies potential dangers of non-taboo inputs (permissible referring expressions) producing taboo outputs. Attempts to predict image offensiveness within the context of an input text are likely to encounter challenges when inputs are underspecified. See Figure~\ref{fig:examples_risks_b}.

\textbf{Mis/dis-information:} For text-to-image systems which aspire to realism, important ethical concerns arise concerning the deliberate or accidental misleading of viewers' beliefs about the world. Misinformation can lead to adopting addictive habits, belief in pseudoscience or in dangerous health or crisis response information, and other harms (see, e.g., \cite{neumann2022justice}). This is especially risky when systems output photorealistic images, and viewers may be more prone to believe fake photorealistic images than readers are to view fake texts. Identifying mis/dis-information concerns in scene description-to-depiction requires comparing the depicted scene with a model of reality in order to identify misalignments and classify them according to risk of harm. However an underspecified input to a scene description-to-depiction system may have one interpretation which is consistent with reality and alternative interpretations which are not. Underspecification hence risks inadvertent misinterpretation of innocuous inputs, potentially leading to misinformation.
See Figure~\ref{fig:examples_risks_c}.

\textbf{Safety:} Since images can convey meaning (\cref{sec:meaning}), they can mislead with potentially harmful consequences. Instruction manuals, road signs, labels, gestures and facial expressions, and many other forms of visual information can lead viewers to take actions in the world which would potentially lead to harm in inappropriate contexts. As with the misinformation risks concerning underspecification outlined above, there is a risk that inadvertent misinterpretation of innocuous inputs could potentially leading to unsafe images in high-risk scenarios.
See Figure~\ref{fig:examples_risks_d}.

In summary, challenges around input ambiguity seem to exacerbate the risks of many potential concerns around text-to-image systems

\section{Paths Forward}


\begin{figure}
    \centering
    \includegraphics[width=0.45\textwidth]{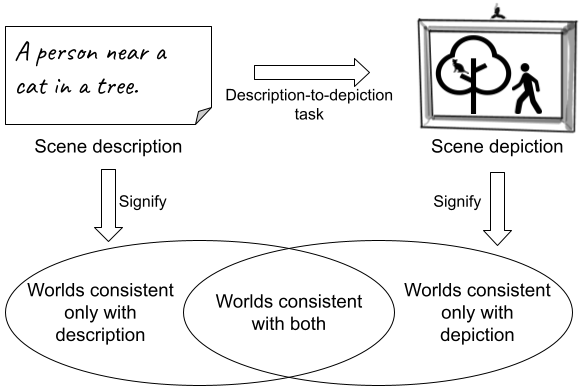}
    \caption{Visual scene depictions and textual scene descriptions may be consistent with different worlds.
    }
    \label{fig:paths_forward}
    \vspace{-0.4cm}
\end{figure}


\subsection{Approaches to input ambiguity} 
 
It is impossible to avoid ambiguous inputs. We describe two possible approaches to managing underspecification in scene-description-to-depiction tasks, which we call \emph{Ambiguity In, Ambiguity Out} (\aiao) and \emph{Ambiguity In, Diversity Out} (\aido). 

The \aiao\  approach posits that a generated image is a model of the intent of the user inputting the text.
As such, this approach proposes that generated depictions should underspecify as much as the input does.
Given the framework in \cref{sec:formulation} whereby scene depictions and descriptions both signify concepts about a (real or imaginary) world fragment, we can consider a depiction $I$ algorithmically generated from a description $T$.
A reasonable assumption regarding image quality is that (all else being equal) the depiction $I$ is better if it is consistent with all and only the same world fragments that $T$ is consistent with. This objective of preserving ambiguity suggests a range of strategies. 
Deliberate visual blurring of non-foreground elements (akin to camera lens and/or exposure effects) can reduce the specificity of objects not mentioned in the text. 
Some visual styles reproduce social stereotypes less than others, for example a stick figure drawing style could minimize depictions of %
phenotypes associated with specific social groups. Orientation choices can be manipulated to obscure information not present in the input text, for example if a figure is facing away from the viewer there may be less need to generate specific facial characteristics.

In contrast, the \aido\ approach acknowledges that since text and image communicate meaning in different ways, it is often extremely challenging or impossible to translate linguistic ambiguities into visual ambiguities (especially discrete structural ambiguities such as PP attachment or word sense ambiguities). This approach instead advocates for systems which output sets of images, such that the diversity of the output set captures the space of interpretations of the input. When asked to depict \textinput{a boss}, the \aido\ approach would aim to show many diverse people. Some challenges that arise include how to measure image diversity in a socially appropriate way \cite{mitchell2020diversity}, as well as what space of possibilities should be represented at all.

Due to the challenges in translating ambiguities between mediums, the \aido\ approach is likely to generally be more tractable and operationalizable in application systems that permit multiple outputs.
However in practice the two approaches are not exclusive and it is possible to combine them. For example, a system generating images for \textinput{a boss} may both generate a set of images that includes both diverse faces (\aido) as well as stick figures and images with obscured facial features (\aiao). Also, the two approaches agree that what is specified in the input should also be specified in the output(s). For example, if asked to depict \textinput{eight tall buildings} then the system should aim to generate an image that provides both perspective and spatial configurations that allow the count of eight buildings to be verified using the image alone.

\subsection{Clarifying tasks and capabilities}
When people collaborate to produce comics, an ``important ingredient is the writer's understanding of the artist's style and capabilities'' \cite{eisner2008comics}---and the same is true of human-machine text-to-image collaborations.
Just as the Bender Rule advocates for explicitly naming the languages of NLP systems \cite{bender2019benderrule}, developers of multimodal systems should aim to understand and communicate the ``visual language'' capabilities of their systems.
Understanding and documenting a deployed text-to-image system's interpretive and generative capabilities---including what visual styles it produces and which text-to-image tasks (\cref{sec:formulation}) it can perform---is therefore important for managing user expectations, aiding users in interpreting system behaviours, and mitigating risks of misuse (\cref{sec:risks}). 
Understanding the landscape of visual capabilities (and also non-capabilities, i.e., both the range and the codomain of the model) will require engaging with experts in visual disciplines, such as photographers, artists, designers, and curators.
We propose that care should be taken when handling training and test data in order to distinguish the semantic and pragmatic relationships between aligned text-image pairs (\cref{sec:relationships}), using relationships which make sense for the tasks and applications at hand.

\subsection{Risk mitigation}

We recommend adopting clear principles of desirable and undesirable system behaviors, especially with regards to biases, offensive and taboo topics, safety, and misinformation risks (\cref{sec:risks}).
Robust stress testing with an adversarial mindset can help to detect corner cases which might trigger undesirable model behaviors, and a culturally diverse pool of stress testers broadens the space of issues which are likely to be detected.
Communicating application-specific uses cases of a text-to-image system \cite[see][]{mitchell2019model} can help to mitigate risk since specific applications come with specific user expectations (e.g., applications for entertainment may not have expectations of truthfulness).  

A description-to-depiction system should take into account the potential effects on viewers concerning sensitive and taboo topics. One simple mitigation strategy is for a system to refuse to generate images which are (predicted to be) harmful or offensive, e.g., based on the offensiveness of the input or analysis of the output. However, even if an image or a text are inoffensive alone, an image can nevertheless be offensive if generated in response to the text; for example neither a portrait of a black woman nor the text \textinput{an angry person} is offensive on their own, yet the former may reproduce the ``angry black woman'' stereotype \cite{walley2009debunking} if generated in response to the latter.

\citet{derczynski2022handling} present recommendations for handling harmful text that are relevant to images. These include using overlays to convey that the contents or associations of the harmful image is not condoned, being transparent about why the image is being used within some context (e.g., as an example of something problematic), stating that the harmful image is harmful, or using cropping, blurring or other visual obfuscation techniques (as adopted, e.g., by \citet{birhane2021multimodal}).


\section{Conclusion}

We have motivated greater consideration of task formulation and underspecification in text-to-image tasks. We laid out the conceptual elements required for this, including greater clarity around the formulation of the space of tasks, as well as consideration of how texts and images each convey concepts. Echoing \citet{van2019pragmatic}, our goal in connecting state-of-the art technologies to theories of cultural and social studies is both to promote deeper understanding of these technologies, and also to foster dialogue across disciplines. We outlined some of the primary challenges concerning textual and visual specification and proposed that systems should consider both reproducing visually the vagueness and ambiguities of the input and producing a diversity of images which convey the breadth of text interpretations. We encourage more work on
measuring visual objectives discussed in cultural fields---such as clarity, aesthetics, etc.---and on
task-specific utility of generated images \cite[cf.][]{fisch2020capwap, zhao2019informative}.

\paragraph{Limitations}
Any position paper at least somewhat reflects the backgrounds and standpoints of its authors. The authors have backgrounds in NLP, computational social science, and AI ethics. Although we call for greater engagement with creative disciplines, we do not represent those disciplines. Although we raise culturally sensitive questions, we have first-hand lived experiences in only Australia, India, the UK and the USA.


\section*{Acknowledgements}
We would like to thank Emily Denton, Kieran Browne, and the anonymous reviewers for their suggestions and feedback.

\bibliography{anthology,custom}
\bibliographystyle{acl_natbib}

\end{document}